\documentclass{article}

\usepackage[preprint]{corl_2026} 
\usepackage[utf8]{inputenc} 
\usepackage[T1]{fontenc}    
\usepackage{hyperref}       
\usepackage{url}            
\usepackage{booktabs}       
\usepackage{amsfonts}       
\usepackage{nicefrac}       
\usepackage{microtype}      
\usepackage{xspace}
\usepackage{amsmath}
\usepackage{amssymb}
\usepackage{graphicx}
\usepackage{subcaption}
\usepackage{cite}
\usepackage{graphicx}   
\usepackage{wrapfig}    
\usepackage{enumitem}
\usepackage[table]{xcolor}
\usepackage{float}

\definecolor{LightRed}{RGB}{227,120,117}
\definecolor{DeepPink}{RGB}{255,20,147}

\newcommand{\ours}{{MemoryWAM}\xspace}

\title{MemoryWAM: Efficient World Action Modeling \\ with Persistent Memory}

\author{\textbf{Sizhe Yang}$^{1,*}$\quad \textbf{Juncheng Mu}$^{2,*}$\quad \textbf{Tianming Wei}$^{2}$\quad \textbf{Chenhao Lu}$^{2}$\quad \textbf{Xiaofan Li}$^{3}$\quad \textbf{Linning Xu}$^{1}$ \\
\textbf{Zhengrong Xue}$^{2}$\quad \textbf{Zhecheng Yuan}$^{2}$\quad \textbf{Dahua Lin}$^{1}$\quad \textbf{Jiangmiao Pang}$^{1}$\quad \textbf{Huazhe Xu}$^{2}$ \vspace{1mm} \\
$^{1}$The Chinese University of Hong Kong
\quad $^{2}$Tsinghua University\quad $^{3}$Zhejiang University \vspace{1mm} \\
\textcolor{gray}{$^{*}$ equal contribution} \vspace{1mm} \\
\textbf{Project page:} \href{https://yangsizhe.github.io/MemoryWAM/}{\textcolor{DeepPink}{https://yangsizhe.github.io/MemoryWAM/}}
}

\begin{document}
\maketitle


\begin{figure}[H]
\centering
\vspace{-5mm}
\includegraphics[width=0.99\columnwidth]{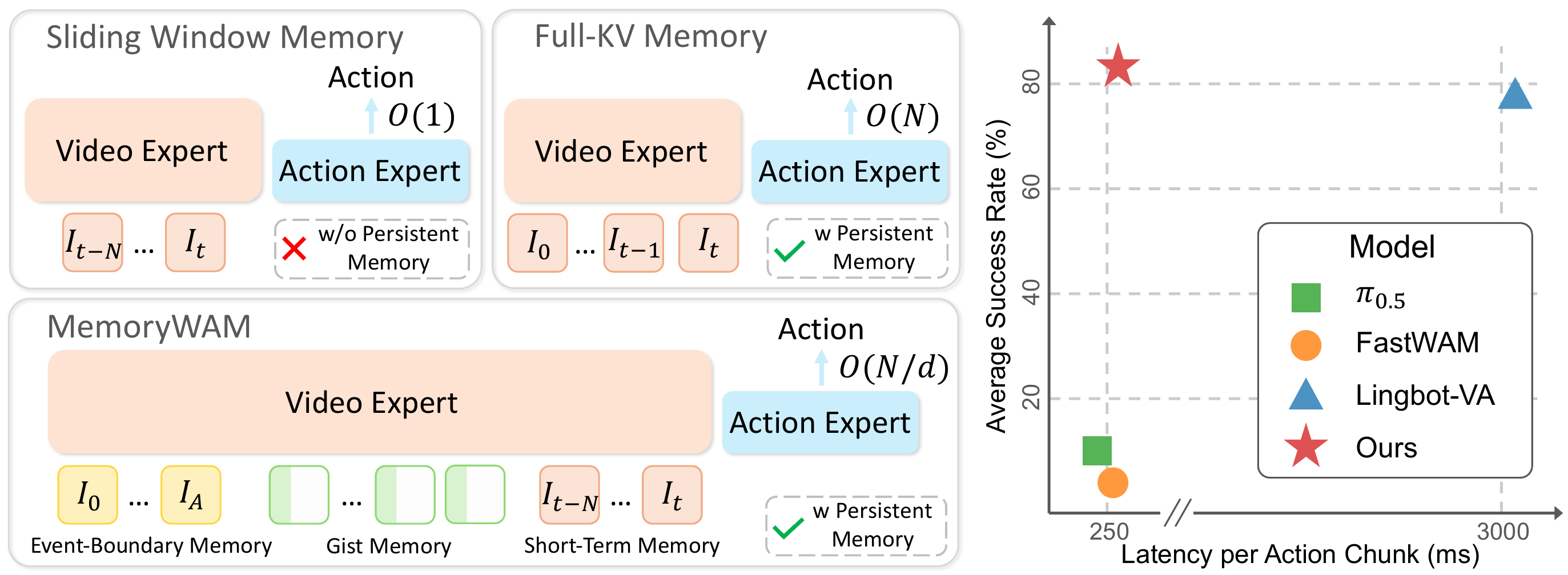}
\caption{\textbf{Overview.} 
Prior WAMs typically face a memory-efficiency trade-off: sliding-window memory is efficient but forgets long-range context, while full-history KV caching preserves context but scales linearly with trajectory length $N$. 
\textbf{\ours} instead introduces \emph{hybrid memory}: recent frames for short-term memory, anchor frames for event-boundary context, and gist tokens for long-range history. This reduces both time and space complexity at inference time from \(O(N)\) to \(O(N/d)\) while preserving persistent context, where $d$ is the compression ratio. On RMBench~\citep{rmbench}, \ours achieves state-of-the-art performance with significantly lower inference latency than full-history WAM baselines.
}
\label{fig:teaser}
\vspace{-2mm}
\end{figure}


\begin{abstract}
Robust robotic manipulation in the real world requires not only an understanding of the current observation, but also \textit{memory} and \textit{dynamics modeling}. 
World action models (WAMs) possess these capabilities by jointly modeling visual foresight and actions conditioned on both current and historical observations, making them a promising paradigm for robotic manipulation. 
However, existing WAMs face a fundamental trade-off: methods with efficient inference typically condition only on a bounded window of recent observations and therefore struggle in non-Markovian environments, whereas methods that preserve long histories incur time and space costs that grow substantially with sequence length. 
To address this challenge, we introduce MemoryWAM, a world action model with efficient persistent memory. 
MemoryWAM uses a hybrid memory design that combines recent frames, event-boundary anchor frames, and compact gist tokens that summarize long-range history. A tailored attention mechanism enables retrieval of both detailed short-term context and compressed long-term context, supporting memory-dependent decision-making with reduced inference latency and GPU memory usage.
Across long-horizon, memory-dependent manipulation tasks in both simulation and the real world, MemoryWAM outperforms strong vision-language-action (VLA) and WAM baselines while maintaining favorable computational efficiency.
\end{abstract}



\section{Introduction}
\label{sec:introduction}

Vision-language-action models (VLAs) have emerged as a dominant paradigm for robotic foundation models, achieving strong generalization by transferring semantic priors from vision language models to robot manipulation~\citep{rt2,openvla,pi0,groot_n1_2025,rdt1b_2024}. 
However, most existing VLAs learn the direct mapping from the current observation to actions. 
While effective for semantically grounded short-horizon skills, they lack memory of historical observations and do not model how the physical world evolves through interaction.
Open-world manipulation instead requires policies to reason not only about \emph{what is visible now}, but also about \emph{what happened before} and \emph{how the environment will evolve}, especially when task-relevant cues are transient, occluded, or have delayed effects. World action models (WAMs) offer this capability by jointly modeling visual foresight and action prediction conditioned on current and historical observations~\citep{lingbot-va,beingh07,DVA,motus,dreamzero}. By grounding manipulation in learned world dynamics, WAMs provide a promising path toward memory-aware, data-efficient robotic manipulation, while also enabling the use of large-scale unlabeled video data beyond costly robot demonstrations.

Despite their promise, existing WAMs face a core memory-efficiency trade-off.
Efficient WAMs such as Cosmos Policy~\citep{cosmos-policy}, DiT4DiT~\citep{dit4dit}, FastWAM~\citep{fastwam}, GigaWorld Policy~\citep{gigaworld-policy}, and X-WAM~\citep{x-wam} condition on a fixed-size window of recent observations. 
Although computationally practical, this design provides only short-term memory and is insufficient for non-Markovian tasks in which crucial information lies outside the current observation window. 
In contrast, autoregressive WAMs such as LingBot-VA~\citep{lingbot-va}, DreamZero~\citep{dreamzero}, and MotuBrain~\citep{motubrain} cache all historical frames as memory. 
Although this strategy preserves richer temporal context, it renders both training and inference inefficient, as latency and memory consumption grow substantially with sequence length.

Cognitive psychology suggests that human memory is not a unitary store, but a hybrid system composed of complementary forms~\citep{atkinson1968human}: 
short-term memory supports ongoing action planning but has limited capacity~\citep{baddeley1974working}; long-term memory tends to preserve abstract gist traces rather than exact verbatim details~\citep{brainerd2005science}; and event boundaries in continuous experience are especially salient for organizing memory~\citep{zacks2007event}.
Inspired by this hybrid organization, we introduce \textbf{\ours}, a world action model with efficient persistent memory. 
\ours implements a hybrid memory mechanism that enables efficient and effective memory utilization for decision-making by using only a small number of gist tokens together with a carefully designed attention mechanism. 
Specifically, a sliding observation window preserves high-fidelity short-term context for immediate control, a small set of gist tokens compresses long-range history, and anchor frames retain complete visual tokens at event boundaries with heightened mnemonic salience, such as the initial observations of a task.

In this way, \ours retains persistent memory while incurring only a slight increase in inference latency and GPU memory consumption as the number of historical frames grows. 
We evaluate \ours against strong baselines on RMBench~\citep{rmbench}, a long-horizon, memory-dependent manipulation benchmark. 
\ours achieves an average success rate approximately 70 percentage points higher than methods that rely only on the current observation or short-term memory, and it even outperforms LingBot-VA~\citep{lingbot-va}, a strong WAM baseline with persistent memory.
Similar advantages are also observed in real-world experiments.
Moreover, \ours achieves substantially lower inference latency and GPU memory usage than previous WAMs with persistent memory. 
Fig.~\ref{fig:teaser} compares different methods in terms of task success rate and inference latency.

In summary, the contributions of this paper are threefold:
\begin{enumerate}
    \renewcommand{\labelenumi}{\arabic{enumi})}

    \item We propose \textbf{\ours}, a world action model with efficient hybrid memory that integrates sliding-window context, gist tokens, and anchor frames to retain persistent history while substantially reducing GPU memory consumption and inference latency.

    \item We present a systematic study of memory mechanisms for world action models, analyzing their trade-offs among inference latency, GPU memory cost, and policy performance.

    \item We show that \ours consistently outperforms strong VLA and WAM baselines on long-horizon, memory-dependent manipulation tasks in both simulation and the real world. 
\end{enumerate}

\section{Related Work}
\label{sec:related_work}

\textbf{Vision-Language-Action Models.}
Vision-language-action (VLA) models have demonstrated strong generalization across diverse tasks and environments by transferring semantic priors from pretrained vision-language models to robotic manipulation~\citep{rt2,octo,openvla,pi0,groot_n1_2025,rdt1b_2024,galaxea_g0_2025,bridgevla,spatialvla,dexvla_2025}. 
Their progress is further supported by large-scale robot datasets~\citep{openx,droid}, enabling policies to acquire broad visuomotor skills from heterogeneous demonstrations. 
However, most VLA approaches remain \emph{policy-centric}: they map observations directly to actions, with temporal structure and physical dynamics only implicitly learned from action-labeled data.
Consequently, physical dynamics are not treated as first-class modeling targets, which may limit data efficiency and robustness in long-horizon manipulation~\citep{wam-robustness}.

\textbf{World Action Models.}
World action models (WAMs) provide a dynamics-centric alternative to direct observation-to-action policies by modeling how the world evolves in conjunction with robot actions.
Early approaches first predict future visual goals and then infer actions from the current observation and the predicted visual goals~\citep{unipi,vidar,gen2act,robodreamer}. 
Other methods move toward unified video-action modeling, in which future observations and actions are learned jointly~\citep{vpp,seer,mimic-video,video-generators-robot-policies,dit4dit,WOG,motubrain,gr2,cosmos-policy,lingbot-va,dreamzero,wallwm}.
Recently, several methods have shown that video prediction can serve primarily as training-time supervision for dynamics modeling, enabling inference without costly video denoising~\citep{fastwam,gigaworld-policy}.
However, most efficient WAMs rely on bounded recent context, whereas full-history methods incur inference latency and GPU memory cost that grow rapidly with sequence length~\citep{lingbot-va,dreamzero}. 
This limitation motivates \ours's efficient persistent memory, which substantially accelerates inference while reducing GPU memory overhead.

\textbf{Memory Mechanisms for Sequential Modeling.}
Memory is central to sequence modeling, motivating mechanisms ranging from recurrent neural networks (RNNs)~\citep{elman1990finding} and long short-term memory (LSTM)~\citep{lstm} to Transformers with full attention~\citep{transformer}, linear attention~\citep{katharopoulos2020transformers,linear-transformer,gated-delta-net}, and test-time training~\citep{ttt}. 
These designs make trade-offs among memory capacity, update efficiency, and retrieval fidelity.
Similar memory-efficiency trade-offs have recently emerged in long-horizon visual systems, including streaming 3D reconstruction~\citep{loger,scal3r,lingbot-map}, long video generation~\citep{frame-pack,context-as-memory,one-minute-video,ttt-done-right}, and memory-dependent robotic manipulation~\citep{mem,rmbench,memoryvla,remem-vla,memer,cronusvla,gated-memory-policy,robomemarena}. 
Our work studies memory for WAMs and introduces a hybrid memory mechanism that retains long-range context while substantially reducing GPU memory overhead and inference latency.


\section{Method}
\label{sec:method}

\subsection{Overview}
Prior approaches in vision-language-action models (VLAs)~\citep{pi0,pi05_2025} and world action models (WAMs)~\citep{beingh07,fastwam} typically map recent observations to actions, relying on a bounded temporal window $o_{t-N:t}$ and task instruction $l$:
\begin{equation}
a_t = \pi_{\text{short}}\big(o_{t-N:t}, l\big).
\end{equation}
While effective for short-horizon tasks, they struggle in non-Markovian environments where decisions depend on long-range history. 

A straightforward approach to retain long-term history is to preserve the full KV cache of all past observations within an autoregressive Transformer \citep{lingbot-va,dreamzero,motubrain}.
While this design allows the policy to access the complete history $o_{1:t}$, it suffers from rapidly increasing GPU memory usage and inference latency as $t$ grows, making it impractical for long-horizon tasks.
To overcome these efficiency limitations, we draw inspiration from human cognition: 
(1) \textbf{short-term memory} supports ongoing action planning but has limited capacity~\citep{baddeley1974working}; 
(2) \textbf{long-term memory} tends to encode abstract \textbf{gist traces} rather than verbatim details~\citep{brainerd2005science}; 
and (3) \textbf{event-boundary memory} emphasizes the state at the onset of a task~\citep{zacks2007event}.
Motivated by these insights, we propose a \emph{hybrid memory} mechanism that preserves high-fidelity short-term context, maintains a small set of gist tokens summarizing long-range history, and retains anchor frames at task onset. 
This human-like hybrid memory significantly reduces inference cost while enabling the policy to reason over long-horizon dependencies.

\begin{figure}
\centering
\includegraphics[width=0.99\columnwidth]{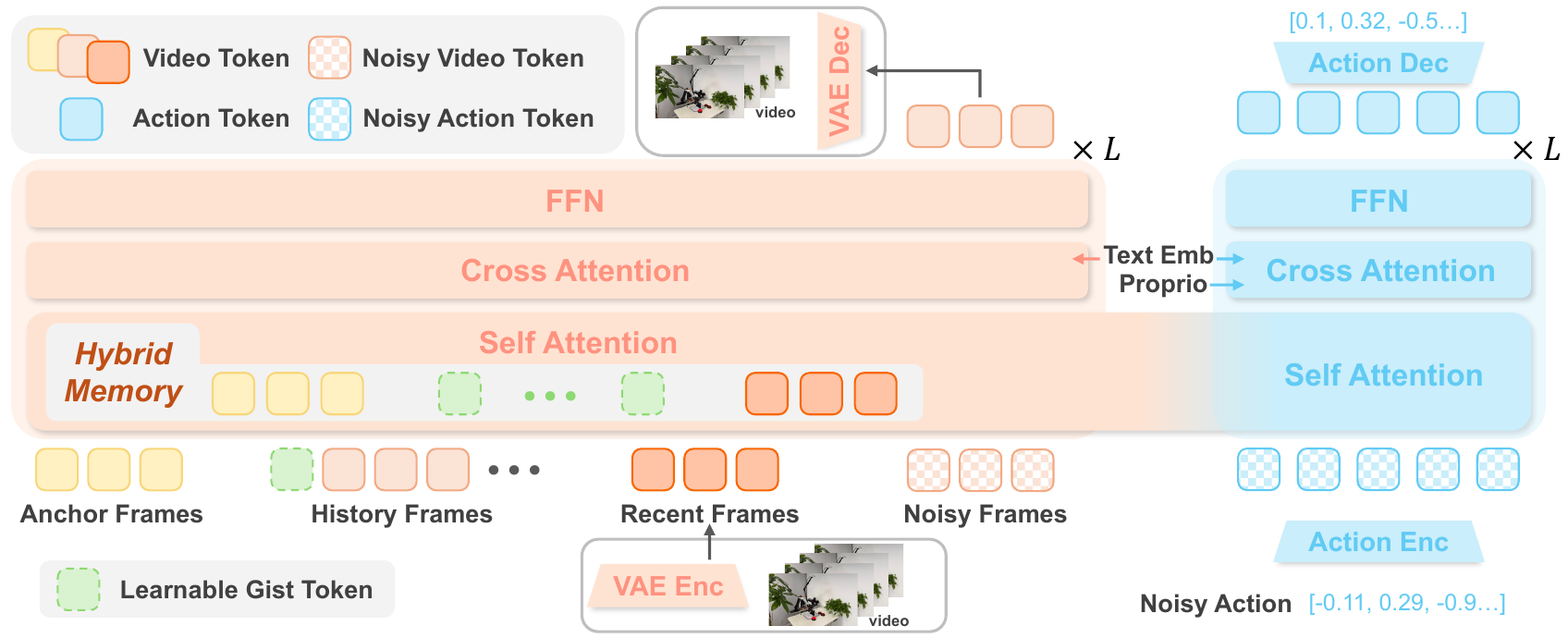}
\caption{
\ours adopts an MoT architecture with a video DiT and an action DiT. 
Video prediction provides dense supervision of dynamics modeling during training and is not required during inference.
For persistent memory, \ours preserves tokens from initial anchor frames and recent frames, and compresses long-range history into a small set of gist tokens. 
This hybrid memory enables non-Markovian decision-making while maintaining low inference latency and GPU memory cost.
}
\vspace{-2mm}
\label{fig:pipeline}
\end{figure}

\subsection{Architecture}
\label{sec:architecture}
\ours follows the recent video-action diffusion paradigm of world action models, where a pretrained video diffusion transformer (DiT) provides dynamics-aware visual representations and a separate action DiT predicts future actions conditioned on the learned visual dynamics. 
The model architecture is illustrated in Fig.~\ref{fig:pipeline}.
Given the observation $o_t$, we first encode it into a compact video latent $z_t$ using a causal video VAE~\citep{wan} for computational efficiency. 
The video latent is processed by a video DiT $\Phi_v$, while action chunks $a_{t:t+h-1}$ are generated by an action DiT $\Phi_a$. 
The two branches are organized in a mixture-of-transformers (MoT)~\citep{mot} architecture.
Moreover, \ours inherits a key advantage of recent efficient WAMs \citep{fastwam,gigaworld-policy}: it learns physical dynamics through video prediction during training, while avoiding expensive video generation at inference time.

During inference, the clean latent $z_t$ of the current observation is forwarded through the video DiT only once to update the video-side key-value (KV) cache $\mathcal{C}_t^v$:
\begin{equation}
    \mathcal{C}_t^v = \Phi_v(z_t, l; \mathcal{C}_{<t}),
\end{equation}
where $\mathcal{C}_{<t}$ denotes the accumulated temporal context. 
The action DiT then predicts the action chunk by denoising action tokens while attending to the cached video representations:
\begin{equation}
    a_{t:t+h-1}
    =
    \Phi_a
    \big(
        x_\tau^a, l;
        \mathcal{C}_{\leq t}^v
    \big),
\end{equation}
where $x_\tau^a$ denotes the noisy action tokens at diffusion time $\tau$. 
Thus, the video DiT extracts dynamics-aware features and maintains memory, while the action DiT maps these features to actions.

\subsection{Hybrid Memory}
\label{sec:hybrid_memory}

As discussed above, full-history attention can lead to rapidly increasing GPU memory cost and inference latency. 
\ours addresses this issue with a hybrid memory design inspired by complementary forms of human memory: \textbf{short-term memory} for immediate closed-loop control, \textbf{event-boundary memory} for retrieval of the state at task onset, and \textbf{gist memory} for compact long-range history.
Formally, at time step $t$, \ours maintains a compact temporal cache,
\begin{equation}
    \mathcal{C}_{\leq t}^{v}
    =
    \mathcal{C}_{\text{short}}^{v}
    \cup
    \mathcal{C}_{\text{anchor}}^{v}
    \cup
    \mathcal{C}_{\text{gist}}^{v},
\end{equation}
where the three components correspond respectively to recent observations, event-boundary frames, and compressed long-term history.

\textbf{Short-term memory.}
Short-term memory is responsible for immediate closed-loop control, where recent observations capture rapidly changing interaction cues such as object motion, contact state, and hand-object configuration. 
We instantiate this memory as a sliding-window cache over the most recent $N_{\text{recent}}$ video frames. 
This preserves high-fidelity local context for action generation, while bounding the short-term attention cost by a constant window size.

\textbf{Event-boundary memory.}
Not all historical observations are equally informative. 
In continuous experience, event boundaries provide salient information for memory organization. 
In robotic manipulation, such boundaries often correspond to task initiation and initial scene configurations. 
We therefore preserve a small set of anchor frames at task onset with full visual tokens, since the initial scene state often grounds key information in the instruction and may later become occluded or fall outside the observation window.

\begin{wrapfigure}{r}{0.5\textwidth} 
    \centering
    \vspace{-8mm}
    \includegraphics[width=\linewidth]{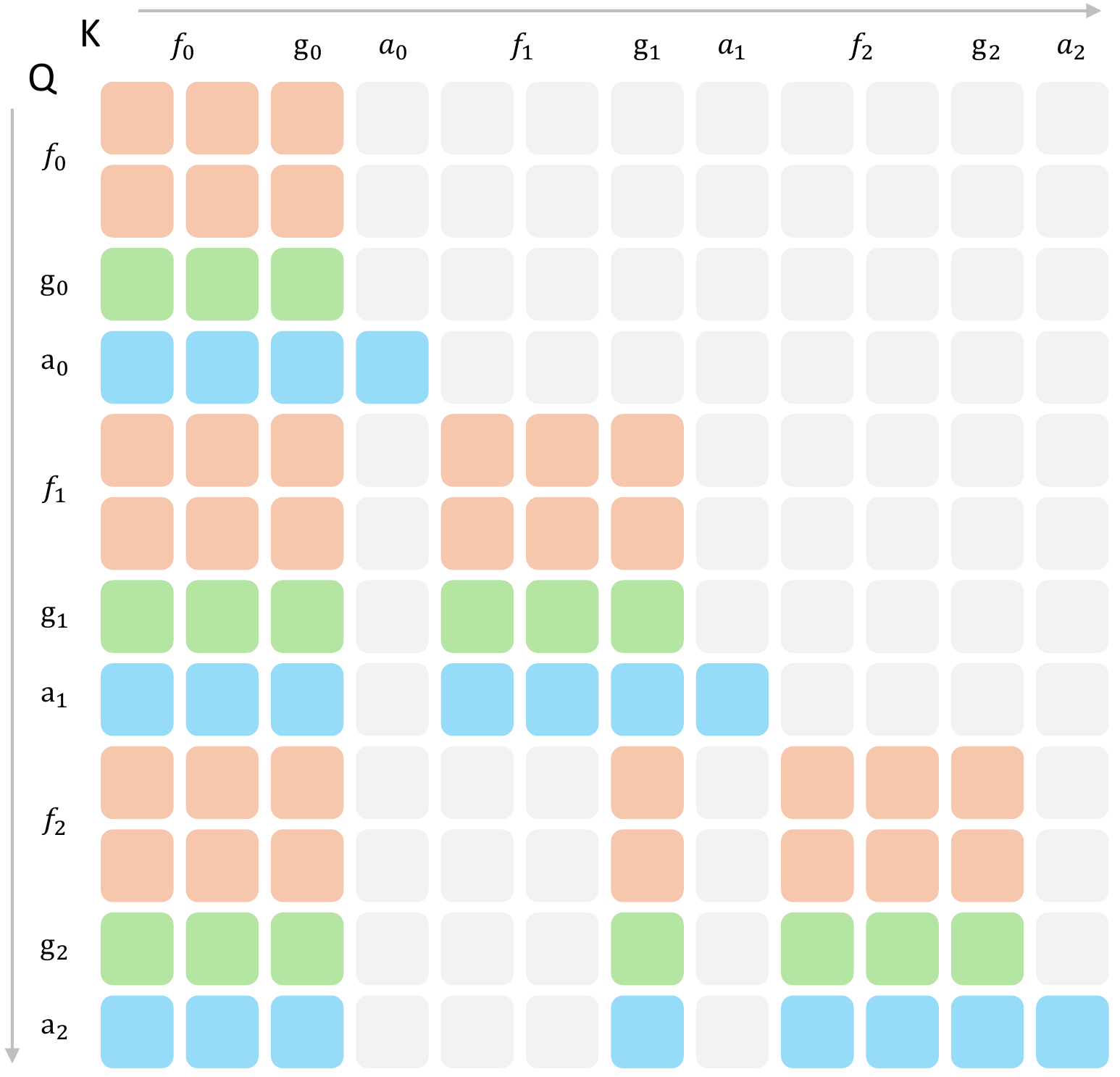}
    \vspace{-5mm}
    \caption{Attention mask of \ours. Example with three frames, one anchor frame, and one recent frame. \(f\) denotes clean video frames, \(g\) indicates gist tokens, and \(a\) represents actions to be denoised. The video frames to be denoised are omitted, as they and the actions attend to the same historical context.}
    \vspace{5mm}
    \label{fig:attention_mask}
\end{wrapfigure}

\textbf{Gist memory.}
While short-term and event-boundary memories preserve selected frames in their entirety, they cannot represent the full long-range history. 
Let each video frame contain $L$ visual tokens. 
Then the number of cached video tokens for full-history attention after $N$ frames is
\begin{equation}
    |\mathcal{C}_{\text{full}}^{v}| = O(NL),
\end{equation}
and both KV-cache storage and attention cost grow linearly with $N$.

To maintain efficient long-term memory, \ours attaches $M$ learnable gist tokens to each frame, where $M \ll L$. 
Given the $L$ visual tokens of frame $f_t$ at time $t$, the corresponding gist tokens $g_t$ attend to both $f_t$ and its historical context.
For a video frame $f_i$ that is neither an anchor frame nor a recent frame, subsequent video and action tokens do not attend to $f_i$ directly; instead, they attend to the corresponding gist tokens $g_i$, which form a compressed representation of $f_i$. 
The attention mask of \ours is illustrated in Fig.~\ref{fig:attention_mask}.
During inference, \ours evicts the KV cache of $f_i$ while preserving the KV cache of $g_i$. 
Consequently, long-range history is retained as a compact persistent memory rather than as a costly full-token KV cache.

This design substantially reduces the time and space complexity of long-term memory. 
If the compression ratio is defined as $d = L/M$, then the long-term cache size becomes
\begin{equation}
    |\mathcal{C}_{\mathrm{gist}}^{v}|
    =
    O(NM)
    =
    O\left(\frac{NL}{d}\right).
\end{equation}
Since $L$ is fixed for a given latent resolution, \ours reduces the sequence-length-dependent storage and attention cost from $O(N)$ to $O(N/d)$ with respect to trajectory length. 
In our implementation, each video frame contains $L=120$ latent visual tokens, while \ours uses only $M=8$ gist tokens per frame, yielding a compression ratio of $d=L/M=15$. 
Thus, for long-term memory, \ours reduces the KV cache by $15\times$ compared with full-history attention.

During action generation, the action DiT attends to the hybrid video cache:
\begin{equation}
    a_{t:t+h-1}
    =
    \Phi_a
    \big(
        x_\tau^a, l;
        \mathcal{C}_{\text{short}}^{v}
        \cup
        \mathcal{C}_{\text{anchor}}^{v}
        \cup
        \mathcal{C}_{\text{gist}}^{v}
    \big).
\end{equation} 
This unified attention interface allows \ours to integrate high-fidelity local context, preserved task-boundary information, and compact long-term history for memory-dependent action generation.

\section{Experiments}
\label{sec:experiments}

\ours aims to equip world action models with efficient persistent memory, enabling end-to-end execution of long-horizon manipulation tasks with low inference cost. 
In this section, we evaluate \ours from three complementary perspectives: efficiency, policy performance, and design effectiveness. 
We first describe the implementation details of \ours, including model architecture, training setup, and inference protocol in Sec.~\ref{sec:implementation_details}. 
We then conduct a systematic study of memory mechanisms in Sec.~\ref{sec:memory_study}, comparing different memory designs in terms of inference latency, GPU memory overhead, and task performance. 
Next, we evaluate \ours on challenging memory-dependent manipulation tasks in simulation (Sec.~\ref{sec:simulation_experiments}) and the real world (Sec.~\ref{sec:real_world_experiments}). 
Finally, we provide comprehensive ablation studies (Sec.~\ref{sec:ablations}) to validate the contribution of each component in our hybrid memory design.

\subsection{Implementation Details}
\label{sec:implementation_details}


\textbf{Model architecture.}
We build \ours on top of the pretrained Wan2.2-TI2V-5B~\citep{wan}, using its video DiT (hidden dim $3072$, FFN dim $14336$, $24$ heads with head dim $128$, $30$ transformer blocks, patch size $1{\times}2{\times}2$ over the
$48$-channel latent), its T5 text encoder, and its 3D causal video VAE. Following FastWAM, the action expert is a separate action DiT that mirrors the video DiT's depth ($30$ blocks) and attention shape ($24$ heads, head dim $128$), but
uses a reduced hidden dimension of $d_a = 1024$ and FFN dim $4096$, yielding a 1B action expert and a total model size of approximately 6B parameters. Following LingBot-VA~\citep{lingbot-va}, we initialize the weights of the action DiT by interpolating the pretrained video DiT along the hidden dimension. The action horizon is set to $h=16$, obtained from a frame stride of $4$ and a temporal VAE stride of $4$, so that each latent frame corresponds to one action chunk of $16$ steps. For RMBench~\citep{rmbench}, images from the
head, left-wrist, and right-wrist cameras are first concatenated into a single $384{\times}320$ mosaic (head at $256{\times}320$ on the bottom; left/right at $128{\times}160$ each, concatenated along the width to $128{\times}320$ on
top) and then jointly encoded by the Wan2.2 VAE, yielding $120$ tokens per video frame after patchification. The robot state and action are both $14$-dimensional joint vectors (dual-arm); proprioception is projected by a learned linear
layer to the text-token dimension and appended to the text context for the action expert. For the hybrid memory module, we keep $M_v=8$ learnable video gist tokens per frame, a sink window of
$N_{\text{init}}=2$ initial frames, and a sliding window of $N_{\text{recent}}=4$ recent clean frames. Gist tokens are realized as learnable parameters and are
placed in the same 3D RoPE coordinate system as their associated video frame, with $(h,w)$ pinned to a constant marker; for the action expert we share the video's 3D RoPE basis so that action queries and cached video keys live in a single positional frame.

\textbf{Training setup.}
We use the same continuous flow-matching formulation for both video and action branches with $1000$ training timesteps. We adopt a shifted logit-normal distribution over $t$ as the noise schedule, with a shift of $5.0$ for the video
branch and $1.0$ for the action branch. For each episode, we build a per-frame autoregressive sequence of interleaved clean/noisy video tokens together with the corresponding action chunks, and apply the hybrid
memory attention mask described in Sec.~\ref{sec:hybrid_memory} so that the training-time visibility exactly reproduces the inference-time KV cache. Following~\citep{selfforcing}, we further augment every clean conditioning latent by linearly mixing it with Gaussian noise at a uniformly random ratio in $[0,1]$ (with probability $p{=}1.0$, applied only on the video side), which prevents teacher-forced training from overfitting to perfectly clean
conditioning frames. We optimize with AdamW (learning rate $2{\times}10^{-4}$, weight decay of $0.01$, $\beta{=}(0.9, 0.95)$) on $8$ GPUs with a per-GPU batch size of $1$. The total loss is $\mathcal{L} = \lambda_v\,\mathcal{L}_{\text{video}} + \lambda_a\,\mathcal{L}_{\text{action}}$ with $\lambda_v = \lambda_a = 1.0$, where each MSE term is reweighted by the scheduler's logit-normal training weight. 

\subsection{Comparison of Memory Mechanisms}
\label{sec:memory_study}

\ours is designed to mitigate the rapidly increasing inference latency and GPU memory usage of full attention. 
To evaluate the effectiveness of the proposed hybrid memory, we compare it with three representative memory mechanisms: full attention~\citep{transformer}, test-time-training (TTT)~\citep{ttt}, and recurrent neural networks (RNNs)~\citep{elman1990finding}. 
The integration of TTT and RNN modules into video diffusion models follows~\citet{ttt-done-right}. Specifically, the original self-attention layer is modified to use sliding-window attention to preserve the capabilities of the pretrained model, while the TTT or RNN module captures long-range temporal dependencies. The outputs of the self-attention layer and the TTT or RNN module are then combined via element-wise addition and used as the input to the next layer.

\begin{figure}
\centering
\vspace{-5mm}
\includegraphics[width=0.99\columnwidth]{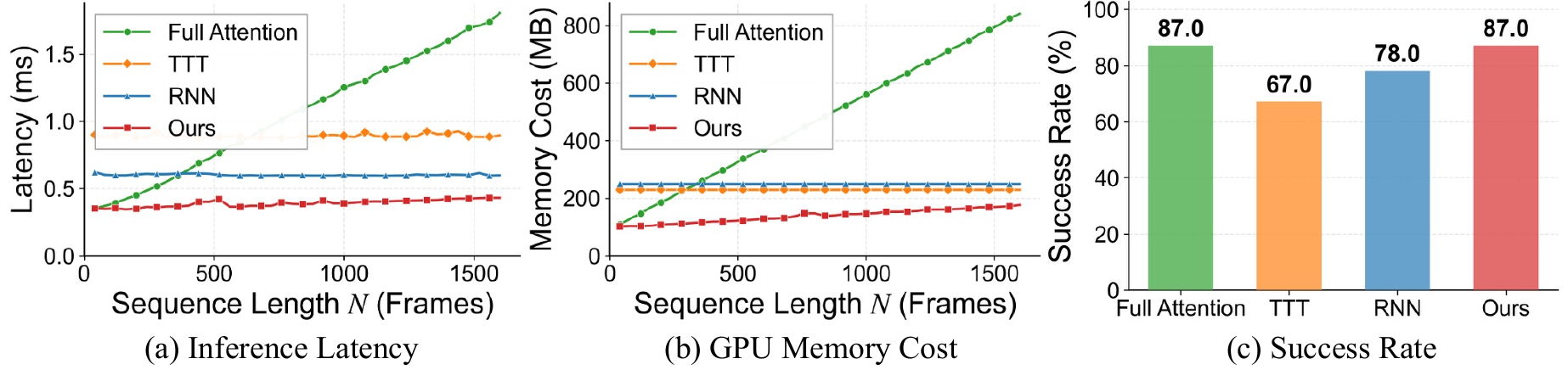}
\vspace{-2mm}
\caption{Comparison of memory mechanisms.
We compare full attention, test-time training (TTT), recurrent neural networks (RNNs), and our hybrid memory mechanism with respect to (a) single-pass inference latency and (b) GPU memory usage as functions of sequence length, and (c) success rates on the Press Button task. Latency and GPU memory usage are measured for a single layer.
}
\vspace{-5mm}
\label{fig:efficiency}
\end{figure}

For a controlled comparison of efficiency, we evaluate all four memory mechanisms within a single layer and measure how their single-pass inference latency and GPU memory usage scale with sequence length, as shown in Fig.~\ref{fig:efficiency}~(a,b). 
TTT- and RNN-based memory maintain constant complexity with respect to sequence length, since they compress history into a fixed-size state or parameters. 
However, they introduce additional network parameters and update operations, leading to relatively high latency and memory usage even for short trajectories. 
Full attention preserves the complete historical KV cache, causing its latency and GPU memory usage to increase rapidly with trajectory length.
Hybrid memory provides a more favorable trade-off: by preserving only initial and recent high-fidelity context and a small set of gist tokens for long-range history, it substantially reduces both inference latency and GPU memory usage. 
Notably, even at a trajectory length of 1,600 frames, hybrid memory remains more efficient than both RNN- and TTT-based alternatives.

We further evaluate the performance of WAM variants equipped with different memory mechanisms on the challenging Press Button task in RMBench~\citep{rmbench}, as shown in Fig.~\ref{fig:efficiency}~(c).
RNN- and TTT-based memory mechanisms achieve lower success rates, suggesting that overly compressed or update-based states struggle to preserve all the task-relevant details required for memory-dependent manipulation. 
Full attention performs strongly and achieves an 87\% success rate by retaining complete historical context, but at a much higher computational cost. 
Our proposed hybrid memory achieves the same 87\% success rate as full attention while being substantially more efficient.
These results demonstrate that the proposed hybrid memory offers an effective balance between long-term context retention, inference efficiency, and downstream manipulation performance.

\subsection{Simulation Experiments}
\label{sec:simulation_experiments}

We evaluate \ours on RMBench~\citep{rmbench}, a challenging simulation benchmark for long-horizon, memory-dependent robotic manipulation. 
Unlike common long-horizon manipulation benchmarks, where task-relevant information is typically available from the current observation, RMBench requires policies to retain and retrieve historical observations, making it well suited for evaluating persistent memory in robotic manipulation. 
RMBench contains nine dual-arm manipulation tasks spanning different levels of Task Memory Complexity. 
Following the benchmark protocol, we train all methods with 50 expert demonstrations per task and report success rates over 100 rollouts.
We compare \ours against competitive VLA and WAM baselines, including $\pi_{0.5}$~\citep{pi05_2025}, FastWAM~\citep{fastwam}, and LingBot-VA~\citep{lingbot-va}. 
These baselines cover three representative paradigms: direct observation-to-action mapping, efficient WAMs with bounded observation windows, and WAMs with full history.

\begin{table}[t]
\vspace{-7mm}
\caption{\textbf{Results on RMBench~\citep{rmbench}.} We report the success rates over 100 rollouts.}
\vspace{2mm}
\centering
\resizebox{0.75\textwidth}{!}{
\begin{tabular}{l|c c c c}
\toprule
\textbf{Task} & $\boldsymbol{\pi_{0.5}}$ & \textbf{FastWAM} & \textbf{Lingbot-VA} & \textbf{Ours} \\
\midrule
Observe and Pick Up & $9\%$ & $0\%$ & $13\%$ & $\textbf{27\%}$ \\
Rearrange Blocks    & $13\%$ & $0\%$ & $\textbf{100\%}$ & $\textbf{100\%}$ \\
Put Back Block      & $11\%$ & $0\%$ & $\textbf{100\%}$ & $\textbf{100\%}$ \\
Swap Blocks         & $24\%$ & $0\%$ & $99\%$ & $\textbf{100\%}$ \\
Swap T              & $15\%$ & $7\%$ & $88\%$ & $\textbf{94\%}$ \\
Battery Try         & $16\%$ & $20\%$ & $\textbf{41\%}$ & $\textbf{41\%}$ \\
Blocks Ranking Try  & $6\%$ & $26\%$ & $\textbf{100\%}$ & $\textbf{100\%}$ \\
Cover Blocks        & $0\%$ & $0\% $ & $79\%$ & $\textbf{98\%}$ \\
Press Button        & $0\%$ & $0\% $ & $84\%$ & $\textbf{87\%}$ \\
\midrule
\textit{\textbf{Average}} & $10.4\%$ & $5.9\%$ & $78.2\%$ & $\textbf{83.0\%}$ \\
\bottomrule
\end{tabular}}
\label{tab:sim}
\vspace{-3mm}
\end{table}

The results are reported in Tab.~\ref{tab:sim}. 
Since RMBench is designed to evaluate non-Markovian decision-making, baselines that rely on a bounded observation window, such as $\pi_{0.5}$ and FastWAM, fail on most tasks, achieving success rates of only 10.4\% and 5.9\%, respectively.
LingBot-VA preserves the full historical KV cache and therefore achieves strong performance on most tasks, confirming the importance of long-term memory. 
\ours further improves the average success rate by 4.8 percentage points over LingBot-VA and achieves leading performance on every task.
This suggests that retaining all historical tokens is not the only effective way to support persistent memory: by preserving full tokens for key observations and compressing long-range history into gist tokens, \ours retains task-relevant context in a more compact form.

\subsection{Real-World Experiments}
\label{sec:real_world_experiments}

\begin{wrapfigure}{r}{0.57\textwidth} 
    \centering
    \vspace{-5mm}
    \includegraphics[width=\linewidth]{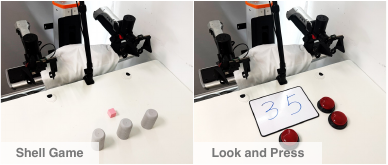}
    \vspace{-5mm}
    \caption{Illustration of the real-world tasks.}
    \vspace{-4mm}
    \label{fig:task_illustration}
\end{wrapfigure}

Our hardware platform consists of an ARX dual-arm robot and a RealSense D455 camera that provides RGB observations. 
We compare \ours with two representative baselines: $\pi_{0.5}$~\citep{pi05_2025} and LingBot-VA~\citep{lingbot-va}. 
We design two challenging memory-dependent tasks, \textit{Shell Game} and \textit{Look and Press}, as shown in Fig.~\ref{fig:task_illustration}.
In \textit{Shell Game}, the robot should identify the cup covering a small cube after a human swaps the cups. 
In \textit{Look and Press}, the robot observes two numbers on the table, presses the left and right buttons the corresponding number of times according to the observed numbers, and finally presses the rear button once to indicate completion.


\begin{wraptable}{r}{0.42\linewidth}
\vspace{-4mm}
\centering
\caption{\textbf{Results of real-world experiments.} We report the number of successes over the total number of trials.}
\vspace{-2mm}
\resizebox{\linewidth}{!}{
\begin{tabular}{l|c c c}
\toprule
\textbf{Task} & $\boldsymbol{\pi_{0.5}}$ & \textbf{Lingbot-VA} & \textbf{Ours} \\
\midrule
Shell Game & $5/20$ & $13/20$ & $\textbf{18/20}$ \\
Look and Press & $0/20$ & $14/20$ & $\textbf{15/20}$ \\
\bottomrule
\end{tabular}
}
\label{tab:real}
\vspace{-3mm}
\end{wraptable}

The results are reported in Tab.~\ref{tab:real}. 
Consistent with the simulation experiments, 
policies with only a short observation window struggle on memory-dependent tasks, while LingBot-VA improves by retaining full history. \ours achieves the best performance on both tasks with substantially lower latency and GPU memory cost than LingBot-VA.
Notably, the high inference latency of LingBot-VA causes it to miss the cup swaps in the Shell Game, leading to task failure. 
These results demonstrate that the proposed hybrid memory not only improves efficiency but also provides a more compact and task-relevant memory representation for real-time, long-horizon robotic manipulation.

\subsection{Effectiveness of Design Choices}
\label{sec:ablations}

To systematically validate the necessity of each component in the proposed hybrid memory, we conduct ablation studies on two challenging tasks from RMBench~\citep{rmbench}, \textit{Cover Blocks} and \textit{Press Button}. 
We compare \ours with four variants: 
(1) \textit{w/o Anchor Frames}, which removes the original video latents corresponding to event-boundary observations from the context and uses gist tokens as a substitute; 
(2) \textit{w/o Gist Tokens}, which removes the long-term gist memory; 
(3) \textit{w/o Sliding Window}, which removes the original video latents corresponding to recent frames from the context and uses gist tokens as a substitute; 
and (4) \textit{Full Attention}, which retains all historical video latents in the context without compression or eviction.

The results are reported in Tab.~\ref{tab:ablation}. 
Different tasks exhibit different sensitivities to memory components, reflecting their distinct temporal dependencies. 
Removing gist tokens causes the largest performance drop, indicating that long-term history is essential for memory-dependent decision-making. 
Removing anchor frames or the sliding window also degrades performance, showing that task-boundary information and high-fidelity recent observations provide complementary benefits. 
Compared with the hybrid memory, full attention retains all historical video latents in the context, but achieves weaker performance. 
This suggests that retaining the entire history is not always optimal: dense historical context can introduce redundant information and make it harder to retrieve task-relevant information. 
Overall, these ablations confirm that \ours's hybrid memory design is not merely an efficiency-oriented compromise, but an effective memory structure that balances short-term observation, event-boundary preservation, and long-term historical abstraction.

\begin{table}[h]
\vspace{-3mm}
\caption{\textbf{Ablation study of the hybrid memory.} We report the success rates of two representative tasks on RMBench.}
\vspace{1mm}
\centering
\resizebox{1.0\textwidth}{!}{
\begin{tabular}{l|c c c c c}
\toprule
\textbf{Task} & \textbf{w/o Anchor Frames} & \textbf{w/o Gist Tokens} & \textbf{w/o Sliding Window}& \textbf{Full Attention} & \textbf{Ours} \\
\midrule
Cover Blocks & $58\%$ & $75\%$ & $96\%$ & $96\%$ & $98\%$ \\
Press Button & $90\%$ & $5\%$ & $69\%$ & $87\%$ & $87\%$ \\
\midrule
\textit{\textbf{Average}} & $74.0\%$ & $40\%$ & $82.5\%$ & $91.5\%$ & $\textbf{92.5\%}$ \\
\bottomrule
\end{tabular}
}
\label{tab:ablation}
\end{table}

\section{Conclusion}
\label{sec:conclusion}

We presented \textbf{\ours}, a world action model with efficient persistent memory for long-horizon robotic manipulation. 
By integrating a sliding observation window, preserved anchor frames, and compact gist tokens, \ours preserves historical context without prohibitive computational cost. 
Across memory-dependent manipulation tasks in both simulation and the real world, \ours outperforms competitive VLA and WAM baselines while achieving practical inference efficiency.

\noindent\textbf{Limitations and future work.} \ours inherits the limitations of video diffusion models, particularly their limited capacity for semantic understanding and reasoning. Future work could address these limitations by incorporating dual-system architectures~\citep{hi-robot, helix} or unified models~\citep{bagel}.




\clearpage


\bibliography{references}  

\clearpage
\appendix
\section{Appendix}

\subsection{Implementation Details}
\label{sup:implementation_details}

\textbf{Training setup.}
Training is conducted in bfloat16 mixed precision with FSDP, activation checkpointing on every DiT block, and gradient clipping at $1.0$.
To ensure a fair comparison with Lingbot-VA, which is pretrained on extensive real-world and simulated data and utilizes action history, we autoregressively incorporate the action history into the action expert for the Swap T task in RMBench, and pretrain our model on the RoboTwin dataset for the Observe and Pick Up task. For all other tasks, however, we neither utilize pretraining nor incorporate the action history. 
Additionally, for the Observe and Pick Up task, we compare the versions without pretraining of \ours and Lingbot-VA, where \ours achieves a 5\% success rate, outperforming Lingbot-VA's 3\%.

\textbf{Inference protocol.}
\ours rolls out autoregressively and maintains, per transformer block, a hybrid memory KV cache consisting of (i) the sink keys/values of the first $N_{\text{init}}=2$ clean video frames, (ii) the keys/values of the
$N_{\text{recent}}=4$ most recent clean video frames (older clean frames are evicted), and (iii) the keys/values of the $M_v=8$ context tokens of every past frame (never evicted). After each action chunk is executed, the new observation mosaics are encoded by the VAE. The corresponding keys and values of the clean latent frames and gist tokens are subsequently prefilled into the cache. Finally, the action expert denoises a new action chunk while attending to the KV cache. We use $50$ flow-matching denoising steps for the action branch in simulation experiments and $10$ denoising steps in real-world experiments. In our
closed-loop control setting, video generation is disabled. After executing the $16$ predicted actions in the environment, we subsample four mosaics at sub-step indices $\{3,7,11,15\}$, append
them to the observation buffer, re-encode with the VAE, and use the resulting latent as the next conditioning latent frame.

\subsection{Real-World Experiment Details}
\label{sup:real_world_experiment_details}

\begin{wrapfigure}{r}{0.5\textwidth} 
    \centering
    \vspace{-8mm}
    \includegraphics[width=\linewidth]{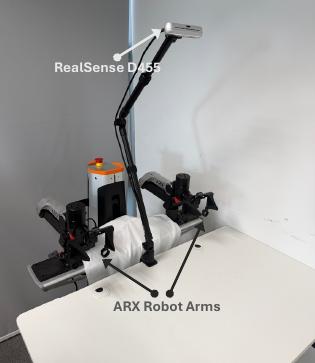}
    \vspace{-5mm}
    \caption{\textbf{Hardware Setup.} The experimental platform comprises a dual-arm robotic system equipped with RealSense D455 cameras for visual perception of the workspace.}
    \vspace{-3mm}
    \label{fig:real_setup}
\end{wrapfigure}

\textbf{Hardware Setup.}
The robotic platform used in the experiments consists of an ARX dual-arm robot, with each arm equipped with a parallel gripper. A RealSense D455 camera captures RGB images of the workspace. The complete hardware setup is illustrated in Fig.~\ref{fig:real_setup}.

\textbf{Imitation Learning Details.}
To validate the application of our method in memory-dependent scenarios, we design two challenging tasks: \textit{Shell Game} and \textit{Look and Press}, as illustrated in Fig.~\ref{fig:task_illustration}.
In the \textit{Shell Game} task, the robot is required to identify and pick up a specific cup that covers a small cube after a human operator randomly swaps the cups. This task is specifically designed to evaluate the policy's ability to track occluded objects over time. For this task, we collect 50 demonstrations.
In the \textit{Look and Press} task, the robot observes two numbers (ranging from 1 to 5) placed on the table. It must then press the left and right buttons a corresponding number of times based on the observed numbers, and finally press a rear button once to indicate task completion. This task assesses the model's counting and working memory capabilities. For this task, we collect 100 demonstrations.
Regarding the hardware and deployment details, the input images captured by the cameras are cropped and resized to a resolution of 
256 × 352. The trained model is deployed on a single NVIDIA RTX 4090 GPU. During real-world execution, the robot operates at a control frequency of 10 Hz within each action chunk. Furthermore, to account for the model's inference time, there is an inter-chunk control latency of approximately 0.3 seconds.

\end{document}